\documentclass[12pt]{iopart}
\usepackage{graphicx}
\usepackage{cleveref}
\usepackage{etoolbox}
\crefname{equation}{}{}
\patchcmd{\numparts}{\addtocounter{equation}{1}}{\refstepcounter{equation}}{}{}
\begin{document}

\title{Constraints on parameter choices for successful reservoir computing}
\author{L. Storm, K. Gustavsson, and B. Mehlig}
\address{
Department of Physics, Gothenburg University,
41296 Gothenburg, Sweden
}

\begin{abstract}
Echo-state networks are simple models of discrete dynamical systems driven by a time series. By selecting network parameters such that the dynamics of the network is contractive, characterized by a negative maximal Lyapunov exponent, the network may synchronize with the driving signal. Exploiting this synchronization, the echo-state network may be trained to autonomously reproduce the input dynamics, enabling time-series prediction. However, while synchronization is a necessary condition for prediction, it is not sufficient. Here, we study what other conditions are necessary for successful time-series prediction. We identify two key parameters for prediction performance, and conduct a parameter sweep to find regions where prediction is successful. These regions differ significantly depending on whether full or partial phase space information about the input is provided to the network during training. We explain how these regions emerge.
\end{abstract}
\section{Introduction}
Many driven dynamical systems can be found in nature and engineering. Reservoir computing has recently become popular to study in this context, as it yields simple models of such dynamical systems. By exploiting signal-driven synchronization, where the dynamics of the reservoir neurons synchronizes with the input time series, a reservoir computer can be trained to reproduce a time series autonomously \cite{pathak2017using,lim2020predicting,lu2017reservoir,kim2021teaching}. A necessary condition for the synchronization to occur is that the dynamics of the reservoir neurons be contractive; a property ensured by the reservoir dynamics having a negative maximal Lyapunov exponent. In reservoir computing literature, the ability to synchronize is referred to as the \textit{echo-state property}, a term coined by Jaeger in his original paper on echo-state networks (ESNs) \cite{jaeger2001echo}, which is the most common realisation of reservoir networks. The maximal Lyapunov exponent has been the focus of study in several papers due to its close connection to the echo-state property \cite{verstraeten2007experimental, massar2013mean,wainrib2016local}. There is some variation in how the maximal Lyapunov exponent has been defined. In \cite{verstraeten2007experimental}, the Lyapunov exponent is defined in the absence of input. However, as the input has been shown to have a contractive effect on the reservoir dynamics when using the commonly employed tanh activation function \cite{massar2013mean}, the maximal Lyapunov exponent defined in the presence of input is more naturally connected to the echo-state property.

While the echo-state property is a necessary condition for the reproduction of a time series, it is not sufficient. The ability for a reservoir network to reproduce a time series has recently been formally connected to time-delay embedding \cite{hart2020embedding}. The result states that the embedding is possible because the neurons in the reservoir represent different time scales of the input time series, creating an internal representation that captures temporal information. In fact, using time delay embedding, it is possible to reproduce a time series with only partial phase space information. By partial phase space information is meant that only a subset of the components of the time series is used when making the prediction of the time series. The connection between the ability to represent several time scales and prediction performance was first observed in \cite{jaeger2001echo,ozturk2007analysis} and has inspired the design heuristic that the reservoir dynamics should be ``rich" in the sense that the different neurons should display a wide range of dynamics that captures different time scales of the time series. However, other results show that the reservoir connections, which allow the reservoir to represent temporal information, can be removed while still maintaining good prediction performance \cite{pyle2021domain,griffith2021essential}. In this case, time delay embedding is not possible. It is clear that such networks cannot reproduced dynamics with only partial phase space information. The distinction between full and partial-information tasks in reservoir computing was made in \cite{lukovsevivcius2009reservoir}, labelled as non-temporal and temporal tasks respectively, but distinctions between how the reservoir should be designed in the two cases were not discussed there.

In this paper, we investigate the differences in parameter dependence when full or partial phase space information is provided to an echo-state network. We begin by showing that, in the limit of large network dimension, and for a given input time series, the maximal Lyapunov exponent depends only on two parameters that combine several tuning parameters, namely the reservoir dimension, the scale of the reservoir connections (here quantified as the variance of the connection weights), the sparsity of the reservoir connectivity matrix, and the dimension and scale of the input. Sweeping the two parameter combinations identified, we study the difference between the regions where reservoir computing is successful for the cases of full and partial information, and explain the shape of these regions. This includes showing why the maximal Lyapunov exponent has a lower boundary in the case of partial information, and how the commonly employed ridge parameter introduces a lower boundary of the input scale for successful reservoir computing. Interestingly, a condition for successful prediction in the partial-information case is shown to imply that the commonly employed \textit{memory capacity} \cite{jaeger2001echo} must be low, implying that maximizing memory capacity is counterproductive when optimizing performance. Additionally, we show that results concerning the sampling rate in time-delay embedding theory \cite{kantz2004nonlinear} can be applied to the case of partial information to improve performance.

The paper is structured as follows: First, we provide some background on the theory of echo-state networks and how their predictive performance is evaluated. In the following section, we derive a mean-field expression for the maximal Lyapunov exponent using random-matrix theory, arriving at the same result as in \cite{massar2013mean}, but extending it to more general input time series rather than Gaussian noise. This is followed by a section where we describe the methods we use. We then present the results for the case of full and partial phase space information. We conclude with a discussion of the results.
\section{Background}
\subsection{Echo-state networks}
The ESN training dynamics for a reservoir with $N$ neurons and an input signal with $n$ components are given by
\begin{numparts}
\begin{eqnarray}
    r_i(t+1) &= g\left(\sum_{j=1}^NA_{ij}r_j(t)+\sum_{\alpha=1}^nW^{(\rm in)}_{i\alpha}u_\alpha(t)\right),\label{eq:train_dyn_a}\\
    v_i(t+1) &= \sum_{j=1}^NW^{(\rm out)}_{ij}f\left(r_j(t+1)\right).\label{eq:train_out}
\end{eqnarray}
\label{eq:train_dyn}
\end{numparts}%
Here $r_i(t)$ is the state of the $i$:th reservoir neuron at time $t$, and $u_\alpha(t)$ is the $\alpha$:th component of the input signal. The matrix $\mathbf{A}$ is the reservoir connection matrix whose entries $A_{ij}$ represent the connection strength between the reservoir nodes, while $\mathbf{W}^{(\rm in)}$ are the connections between the input and the reservoir. $g(\cdot)$ is the activation function, and $f(\cdot)$ is applied to the reservoir states before it is projected to the output space with the output weight matrix $\mathbf{W}^{(\rm out)}$. The argument of the activation function is referred to as the local field. $f(\cdot)$ is often set to be the identity function. In this work, to break the inherent symmetry of the reservoir dynamics which causes the ESN to learn the reflected input series $\mathbf{u}\rightarrow -\mathbf{u}$ as well as the original, we employ the Lu readout \cite{lu2017reservoir}

During prediction, we follow the standard procedure introduced in \cite{jaeger2001echo} and replace the input by the output of the reservoir to form an autonomous system,
\begin{numparts}
\begin{eqnarray}
    r_i(t+1) &=& g\left(\sum_{j=1}^NA_{ij}r_j(t)+\sum_{\alpha=1}^nW_{i\alpha}^{(\rm in)}v_\alpha(t)\right),\\
    v_i(t+1) &=& \sum_{j=1}^NW^{(\rm out)}_{ij}f\left(r_j(t+1)\right).\label{eq:test_dyn_b}
\end{eqnarray}
\label{eq:test_dyn}%
\end{numparts}
This is the prediction dynamics.
\subsection{Training and evaluation}\label{sec:train_and_eval}
In order to train the ESN, the training dynamics \eref{eq:train_dyn} is run for some time using the input time series to ensure that the reservoir dynamics has synchronized with the input. Then, at time $t=0$, an $N\times T_{max}$ matrix $\mathbf{R}$ is formed where each column is the reservoir state $\mathbf{r}(t)$ at each time $t=0,\;1,\dots, T_{max}-1$. We wish to minimize the quadratic error between the output $\mathbf{v}(t)$ and the target $\mathbf{y}(t)=\mathbf{u}(t)$ and achieve this by employing ridge regression \cite{tikhonov1977solutions} to obtain
\begin{equation}
    \mathbf{W}^{(\rm out)}=\mathbf{y}\mathbf{R}^{\top}(\mathbf{R}\mathbf{R}^{\top}+k\mathbf{I})^{-1}.
    \label{eq:ridge}
\end{equation}
Here, $k\geq0$ is the ridge parameter which is introduced to reduce overfitting. An additional effect of the ridge parameter is that the magnitude of the entries in $\mathbf{W}^{(\rm out)}$ decreases as $k$ increases.

Once $\mathbf{W}^{(\rm out)}$ has been determined, the prediction dynamics \eref{eq:test_dyn} is used to autonomously predict how the time series continues. In order to evaluate the performance of the ESN, we monitor
\begin{equation}
    \varepsilon_\alpha(t)=\sqrt{\frac{\left(y_\alpha(t)-v_\alpha(t)\right)^2}{\sigma_{y_\alpha}^2}},
\end{equation}
where $\sigma_{y_\alpha}^2$ is the variance of the $\alpha$:th component of the time series.  The quantity $\varepsilon_\alpha(t)$ quantifies how many standard deviations the $\alpha$:th component of the prediction deviates from the target time series. When any of the predicted components deviates more than some threshold value, the time is recorded as the successful prediction time. We set the threshold value to 0.5. Decreasing this value does not qualitatively affect the obtained results. As this quantity fluctuates depending on the random initialisation of the ESN and from where in the time series the prediction started, the final performance score is determined by an average over several random initialisations of both the ESN and initial value of the time series. As the quantity is standardized, the metric is comparable for different time series.
\subsection{Parameters}
In designing an ESN, several parameters must be selected. As they are central to this work, we summarise the relevant parameters here. The parameters that are mainly discussed in literature are the reservoir dimension $N$, the scale of the reservoir connectivity matrix $\sigma_A^2$, which is the variance of the entries in $\mathbf{A}$ (the spectral radius is sometimes used instead as a scale metric), the sparsity of the connections in the reservoir $s$, which takes the value $s=1$ if all neurons are connected and $s=0$ if no neurons are connected, the input dimension $n$, and the scale of the input $\sigma_{in}^2$, which is the variance of the entries of $\mathbf{W}^{(\rm in)}$. These are parameters pertaining to the architecture of the ESN. In addition, the ridge parameter $k$ used during training and the sampling rate $\delta t$ of the time series are important tuning parameters.
\section{Maximal Lyapunov exponent}
The maximal Lyapunov exponent of a dynamical system describes the long term fate of the separation of two initially nearby trajectories \cite{ott2002chaos}. The quantity is computed under the assumption that the separation remains small within the time frame of interest, and as such, we can consider the linearised dynamics of the system to describe the evolution of the separation. For echo-state networks, it is possible to define three different Lyapunov exponents by considering different dynamical systems: (i) system \eref{eq:train_dyn_a} with $\sigma_{in}^2=0$, (ii) system \eref{eq:train_dyn_a} with $\sigma_{in}^2>0$, and (iii) system \eref{eq:test_dyn} for a trained ESN. In \cite{verstraeten2007experimental}, definition (i) was employed. However, 
definition (ii) must be used if one wants to quantify the echo-state property, because the input has a contracting effect on the reservoir dynamics when the tanh activation function is employed \cite{massar2013mean}. It is therefore more natural to study the latter definition. Finally, if an ESN has been trained successfully, the third definition of the exponent approximate the maximal Lyapunov exponent of the input dynamics, as shown in \cite{pathak2017using}. We mainly focus on definition (ii) and refer to this as the training Lyapunov exponent $\lambda_T$.

For an ESN employing the tanh activation function, we may compute the linearised separation of reservoir states $\delta\mathbf{r}(t)$ in the presence of input as
\begin{equation}
    \delta\mathbf{r}(t+1)=\mathbf{D}(t)\mathbf{A}\delta\mathbf{r}(t),
\end{equation}
where $\mathbf{D}(t)$ is a diagonal matrix with entries $D_{ii}(t)=1-\tanh^2{\left(b_i(t)\right)}$, where $b_i(t)=\sum_j^NA_{ij}r_j(t)+\sum_\alpha^n W_{i\alpha}^{(\rm in)}u_\alpha(t)$. The training Lyapunov exponent is obtained by computing \cite{ott2002chaos}
\begin{equation}
    \lambda_T=\lim_{t\rightarrow\infty}\frac{1}{t}\log\frac{|\mathbf{D}(t-1)\mathbf{A}\mathbf{D}(t-2)\mathbf{A}\dots\mathbf{D}(0)\mathbf{A}\delta \mathbf{r}(0)|}{|\delta\mathbf{r}(0)|}.
    \label{eq:lya}
\end{equation}
Numerically, the product in \eref{eq:lya} can be computed employing the QR method \cite{geist1990comparison} and computing the average maximal expansion of $\delta\mathbf{r}(t)$ per time step until the average has converged to some fixed value.

The training Lyapunov exponent has previously been derived in the limit of large $N$ using mean-field theory \cite{massar2013mean}. It was assumed that the reservoir dimension $N$ is sufficiently large so that the sum $\sum_{j=1}^NA_{ij}r_j(t)$ is distributed according to a normal distribution due to the central limit theorem. We employ the same assumption and derive a similar result for the training Lyapunov exponent using random matrix theory. We do not assume that the input is Gaussian random noise, but that it is a general, stationary time series with a rapid decay of time correlations. Using these assumptions, we obtain an expression for the training Lyapunov exponent (see Appendix):
\begin{equation}
    \lambda_T=\frac{1}{2}\left[\ln{\left(sN\sigma_A^2\right)}+\ln{\left(N^{-1}\sum_i^N\langle D_{ii}^2(t)\rangle\right)}\right].
    \label{eq:final_lya}
\end{equation}
This is the same result as \cite{massar2013mean}, for relaxed assumptions on the input time series.
To obtain $\langle D_{ii}^2\rangle$, we use the same procedure as \cite{massar2013mean} and construct an iterative map for the variance of the reservoir states $r_i(t)$. Assuming that $N$ is large enough so that the sum $\sum_{j=1}^NA_{ij}r_{j}$ is normally distributed, we can compute the probability density function $f_b(x)$ of the local field by using the convolution of the probability mass function of a normal distribution with zero mean and variance $sN\sigma_A^2\sigma_r^2$, and the empirical probability mass function of the normalized input time series scaled by $\sigma_{in}^2$, to construct an iterative map of the variance of $r_i(t)$,
\begin{equation}
    \sigma_r^2(t+1)=\int_{-\infty}^{\infty}\textrm{d}b\,\left(g(b)\right)^2f_b(b;sN\sigma_A^2,\sigma_r^2(t),\sigma_{in}^2).
\end{equation}
In \cite{massar2013mean}, it was shown that this map converges to a fixed point when the input is a Gaussian random variable. A similar result was derived by Poole et al. \cite{poole2016exponential} for feed-forward neural networks, where the map was also shown to rapidly converge. Our numerical results show that this map also converges for non-Gaussian inputs. Assuming the distribution of $r_i(t)$ has converged to have variance $(\sigma_r^*)^2$, one finds 
\begin{equation}
    \langle D_{ii}^2\rangle=\langle(1-r_i^2(t))^2\rangle=1-2(\sigma_r^*)^2+\langle r_i^4\rangle,
    \label{eq:d_var}
\end{equation}
where the fourth moment of $r_i(t)$, which also converges as the distribution only depends the first and second moments, can be computed as
\begin{equation}
    \langle r_i^4\rangle=\int_{-\infty}^{\infty}\textrm{d}b\,\left(g(b)\right)^4f_b(b;sN\sigma_A^2,(\sigma_r^*)^2,\sigma_{in}^2).
\end{equation}
Combining \eref{eq:final_lya} and \eref{eq:d_var}, we find that the predicted training Lyapunov exponent agrees very well with the result obtained using the QR method when the reservoir dimension $N$ is large. The result shows that $\lambda_T$, for a given input time series, depends on $sN\sigma_A^2$ and $n\sigma_{in}^2$. It is therefore unnecessary to vary $s$, $N$, and $\sigma_A^2$ independently when selecting reservoir parameters, which is often done in literature, see for example \cite{jaeger2001echo, lukovsevivcius2009reservoir}. In the remainder of the article, these two parameters are used to investigate parameter regions where reservoir computing is successful.

\section{Method}
To evaluate the prediction performance of ESNs when full and partial information is provided, we use the ESN to predict a chaotic time series where we either input the ESN with the time series of all the components of the time series, or only a single component. In the latter case, we use the ESN to predict the input component. As the ESN has incomplete information for this case, it must construct a time-delay embedding to reproduce the dynamics correctly. As examples of chaotic time series, we use the Lorenz63 system \cite{lorenz1963deterministic}, given by
\begin{numparts}
\begin{eqnarray}
    \frac{\textrm{d}}{\textrm{d}t}x&=&-\sigma(x+y)\\
    \frac{\textrm{d}}{\textrm{d}t}y&=&\rho x-y-xy\\
    \frac{\textrm{d}}{\textrm{d}t}z&=&xy-\beta z,
\end{eqnarray}
\label{eq:lorenz}
\end{numparts}%
with $\sigma=10$, $\rho=28$, and $\beta=8/3$, which results in that the dynamical system has a Lyapunov spectrum of $\lambda_1=0.901$, $\lambda_2=0$, and $\lambda_3=-14.6$ \cite{sprott2010elegant}, and the Halvorsen system \cite{sprott2010elegant}
\begin{numparts}
\begin{eqnarray}
    \frac{\textrm{d}}{\textrm{d}t}x&=&-ax-4(y+z)-y^2\\
    \frac{\textrm{d}}{\textrm{d}t}y&=&-ay-4(x+z)-z^2\\
    \frac{\textrm{d}}{\textrm{d}t}z&=&-az-4(x+y)-x^2,
\end{eqnarray}
\label{eq:halvor}
\end{numparts}%
with $a=1.3$. The Lyapunov spectrum of the Halvorsen system is $\lambda_1=0.69$, $\lambda_2=0$, and $\lambda_3=-4.9$ when the considered parameters are used \cite{sprott2010elegant}.

We obtain a time series by discretizing the dynamical systems \eref{eq:lorenz} and \eref{eq:halvor} with a sampling rate $\delta t=0.1$. This choice is informed by the work of Kantz and Schreiber (see p. 151 in  \cite{kantz2004nonlinear}) where the information theoretical concept of mutual information is used to find an optimal step size for time delay embedding of the Lorenz63 system. We use the same sampling rate for the Halvorsen time series. The effect of changing the sampling rate is investigated in Section \ref{sec:sampling_rate}. The ESN is trained on the Lorenz63 or Halvorsen system for roughly 200 Lyapunov times. Before feeding the time series to the reservoir, the time series is normalized such that the largest variance of any variable of the dynamical system over time equals unity. This is to ensure that the dependence on $n\sigma_{in}^2$ is comparable for the different time series.
\section{Results and discussion}
\subsection{Parameter dependence for full and partial information}\label{sec:phase_diagram}
\begin{figure}[h]
    \centering
    \includegraphics[scale=0.5]{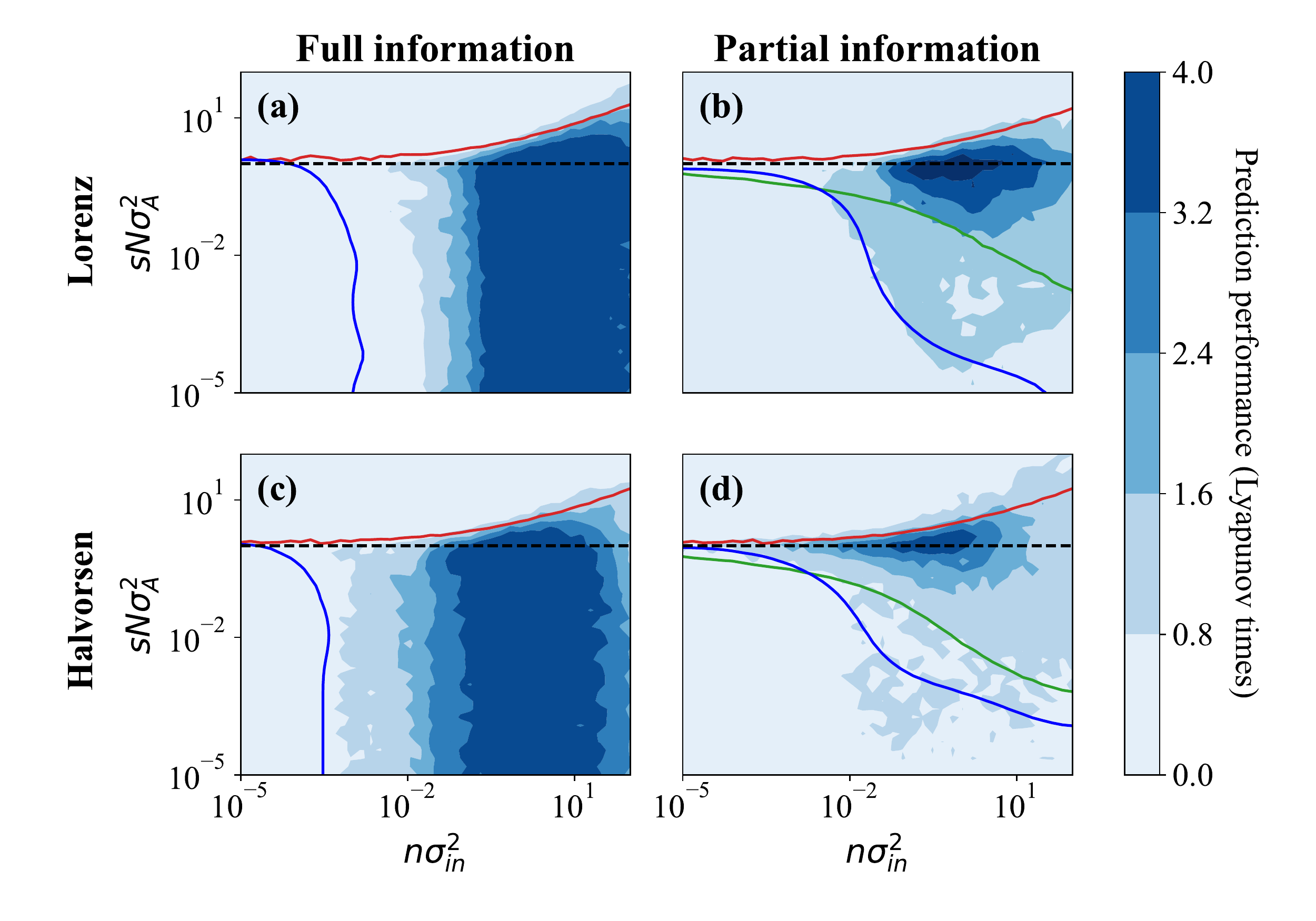}
    \caption{Prediction performance, measured in Lyapunov times, of an ESN of dimension $N=500$, trained on \textbf{(a)} the Lorenz63 system provided with all three components, \textbf{(b)} the Lorenz63 system provided with only the $y$-component, \textbf{(c)} the Halvorsen system provided with three components, and \textbf{(d)} the Halvorsen system provided with only the $y$-component. Each result has been averaged over 50 independent trials. The red line shows where the mean field theory predicts $\lambda_T=0$, while the green line shows where $\textrm{rank}(\mathbf{RR}^\top)=100$. The latter choice is discussed in Section \ref{sec:rank}. The dashed line shows where the maximal Lyapunov exponent in the absence of input is zero, and the blue line shows where the prediction dynamics \eref{eq:test_dyn} bifurcates from a stable fixed point when the ridge parameter $k=10^{-2}$, which was used during training. The results were averaged over 50 independent trials.}
    \label{fig:heatmap}
\end{figure}

We characterize the prediction performance in a phase diagram with axes $sN\sigma_A^2$ and $n\sigma_{in}^2$ (see figure \ref{fig:heatmap}), for two cases: (i) Providing full phase space information to the reservoir (panels \textbf{(a, c)} in figure \ref{fig:heatmap}) and (ii) providing only partial phase space information to the reservoir (panels \textbf{(b, d)} in figure \ref{fig:heatmap}). Different aspects of the phase diagram in figure \ref{fig:heatmap} are discussed below.
\subsubsection{Maximal Lyapunov exponent}
We first observe that the reservoir dynamics must contract ($\lambda_T<0$) for successful prediction. This is demonstrated by the red line in the phase diagrams. In \cite{schrauwen2008computational}, the transition between the successful and failed prediction is shown to be smooth. However, we find that the transition becomes sharper as $N$ increases. We also note that the maximal Lyapunov exponent computed in the absence of input (dashed black line in figure \ref{fig:heatmap}), used in \cite{verstraeten2007experimental}, works well as long as $n\sigma_{in}^2$ is small. As $n\sigma_{in}^2$ becomes larger, the input variance has an increasingly contractive effect on $\lambda_T$. It is clear from figure \ref{fig:heatmap} that $\lambda_T<0$ is a necessary but not sufficient condition for successful prediction.  
\subsubsection{Full and partial information}
A qualitative difference exists in the parameter dependence on prediction performance when full or partial information is provided to the network. In the full information case, as long as $\lambda_T<0$, the performance is roughly independent of $sN\sigma_A^2$. This is consistent with the result of \cite{griffith2021essential} and \cite{pyle2021domain}, where it was shown that the connections between the reservoir neurons can be removed (setting $\mathbf{A}$ to zero) and still the reservoir allows successful prediction. Removing the connections renders the ESN memory-less, and the algorithm simply projects the input series nonlinearly to a high dimensional space and performs a function fitting. This is possible because full phase space information is provided; only the current phase space coordinate is necessary to determine the evolution of the dynamics. This is not the case for partial information. In \cite{hart2020embedding}, it was shown that the reservoir computer employs time delay embedding to predict a time series. It is possible, according to Takens' embedding theorem, to embed a high dimensional time series using the history of a single observable. The theorem states that, given at least $2d_{f}+1$ delays, where $d_{f}$ is the box-counting dimension of the attractor of the time series, the embedding is possible. In our case, this corresponds to having at least $2d_f+1$ neurons representing different time scales of the input time series. The box-counting dimension of the Lorenz63 system is 2.06 \cite{sprott2010elegant}, implying that approximately five neurons are required. However, as was pointed out in \cite{hart2020embedding}, while the embedding is possible, projecting the embedding back to the original space \textit{linearly} \eref{eq:test_dyn_b} is not necessarily accurate. To resolve this, the universal approximation theorem was evoked in \cite{hart2020embedding}, stating that with a sufficiently large sum of weighted nonlinear activation functions, any functional relationship can be approximated. Hence, we need sufficiently many neurons representing different time scales of the input time series to be able to predict the time series when only partial information is provided.
\subsubsection{Rank of $\mathbf{RR}^\top$}\label{sec:rank}
In panels \textbf{(b)} and \textbf{(d)} in figure \ref{fig:heatmap}, the ESN must use time-delay embedding to reconstruct the input dynamics. When $sN\sigma_A^2\sigma_{r}^2\ll n\sigma_{in}^2$, all reservoir states are highly correlated because they are all strongly driven by the input signal. As $sN\sigma_A^2\sigma_{r}^2\sim n\sigma_{in}^2$, the reservoir states may develop different dynamics due to the randomly sampled connections in $\mathbf{A}$. This can be quantified using the rank of the matrix $\mathbf{RR}^\top$, i.e. the number of independent reservoir neurons. We remind the reader that $\mathbf{R}$ is the matrix whose columns are the reservoir states $\mathbf{r}(t)$ throughout the training sequence (see section \ref{sec:train_and_eval}). The rank of $\mathbf{RR}^\top$ quantifies the ``richness" described by Jaeger in his original paper on ESNs. This is the effective number of activation functions that the ESN can use to approximate the functional relationship between the reservoir embedding and the original space. In figure \ref{fig:heatmap}, the green line shows where the rank is equal to 100. Along this contour, the ESN can effectively employ 100 reservoir states to approximate the functional relationship between the time-delay embedding performed by the reservoir and the output. Above the green line, the rank increases gradually, making the approximation more accurate. As shown in figure \ref{fig:heatmap}, it is only once the rank begins to increase that the reservoir is able to predict. The gradual increase of rank is reflected in a gradual increase of performance. In panels \textbf{(a)} and \textbf{(c)}, the rank of $\mathbf{RR}^\top$ does not affect performance, because the ESN does not need to perform a time-delay embedding to reconstruct the input dynamics.

\begin{figure}
    \centering
    \includegraphics[scale=0.5]{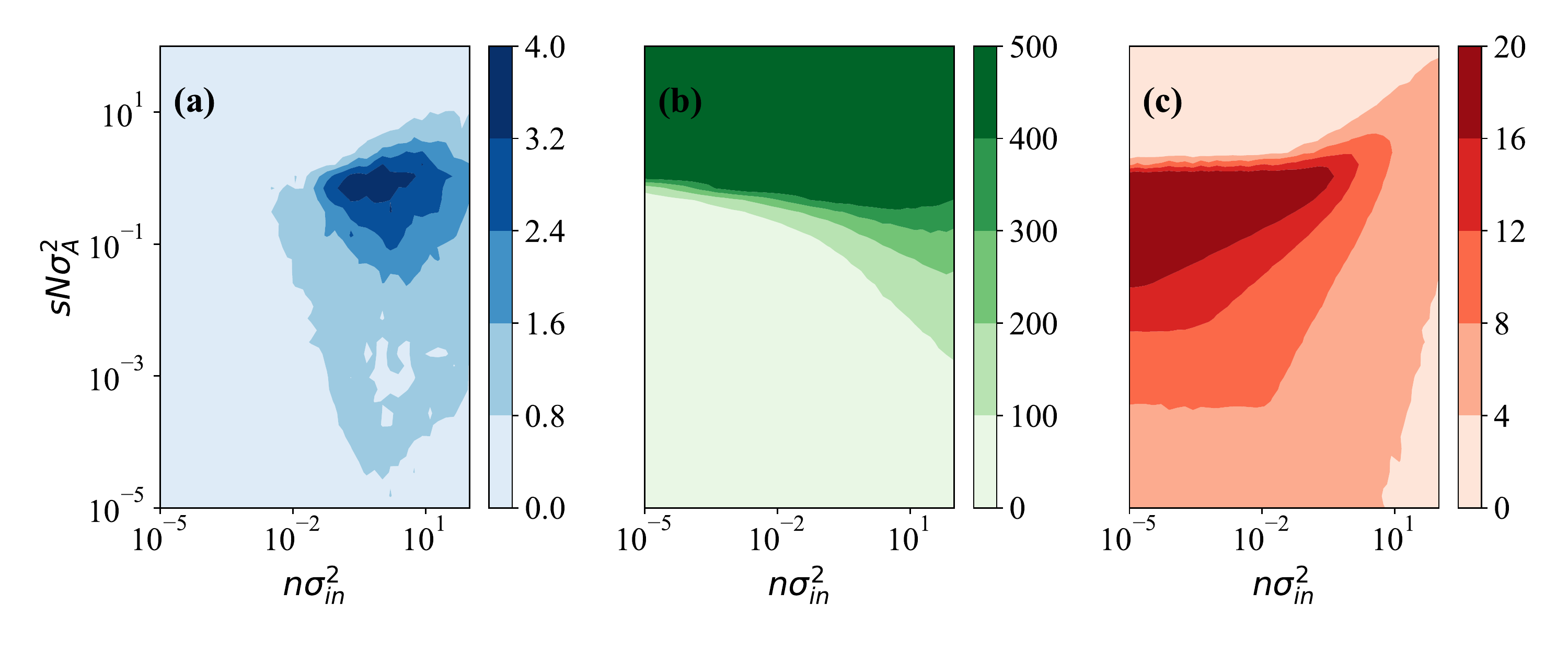}
    \caption{\textbf{(a)} Performance of an ESN with dimension $N=500$ predicting the $y$-component of the Lorenz63 attractor (same data as in figure \ref{fig:heatmap}a), \textbf{(b)} rank of the $\mathbf{RR}^\top$ matrix computed for the same reservoir computer, and \textbf{(c)} the memory capacity, computed for the same reservoir dimension. The result has been averaged over 100 independent trials.}
    \label{fig:mc_rrt_comparison}
\end{figure}

That predictive performance depends on the rank of $\mathbf{RR}^\top$ has several consequences. Firstly, the lower bound is independent on any time scale of the predicted time series. Thus, it is incorrect to state that the scale of $\mathbf{A}$ (often the spectral radius is used) must be adjusted in accordance with the time scale of the predicted time series \cite{jaeger2001echo}. In fact, as long as sufficiently many neurons are uncorrelated and each neuron is an echo of the input, prediction is possible. Secondly, the result has a surprising consequence for the memory capacity of a reservoir \cite{jaeger2001echo}. The memory capacity $MC$ roughly measures how well a reservoir remembers previous inputs and is defined as
\begin{eqnarray}
    MC=\sum_{\tau=1}^\infty\max_{\mathbf{W}^{(\rm out)}}\frac{\textrm{cov}^2(\mathbf{v}(t),\mathbf{u}(t-\tau))}{\sigma_{v_\tau}^2\sigma_u^2},
\end{eqnarray}
where the input is a series of i.i.d. Gaussian random variables. A high memory capacity means that the reservoir state $\mathbf{r}(t)$ contains information about an input $\mathbf{u}(t-\tau)$ for some large $\tau$. Hence, all reservoir states between $t-\tau$ and $\tau$ should be highly correlated. The rank of $\mathbf{RR}^\top$ is equal to its number of non-zero singular values. This is equivalent to the number of non-zero singular values of $\mathbf{R}^\top\mathbf{R}$, which represents the correlations between reservoir states at different times. Since a high rank is needed for good performance, and a low rank reflects a high memory capacity, optimised prediction performance and optimised memory capacity appear to be mutually exclusive. This prediction is verified by figure \ref{fig:mc_rrt_comparison}. Comparing panels \textbf{(b)} and \textbf{(c)}, we see that when the memory capacity peaks, the rank is low. Comparing panels \textbf{(a)} and \textbf{(c)}, we conclude that high memory capacity is not indicative of high prediction performance. This means that prediction performance does not rely on being able to reconstruct the time series far back in time, but rather on the ability to represent several time scales of the input.
\subsubsection{Saturation of activation function}
The performance drops once $n\sigma_{in}^2$ becomes too large. In this limit, the local fields of the reservoir neurons become so large that the activation function saturates and information about the input time series is lost.
\subsubsection{Ridge parameter}
\begin{figure}
    \centering
    \includegraphics[scale=0.6]{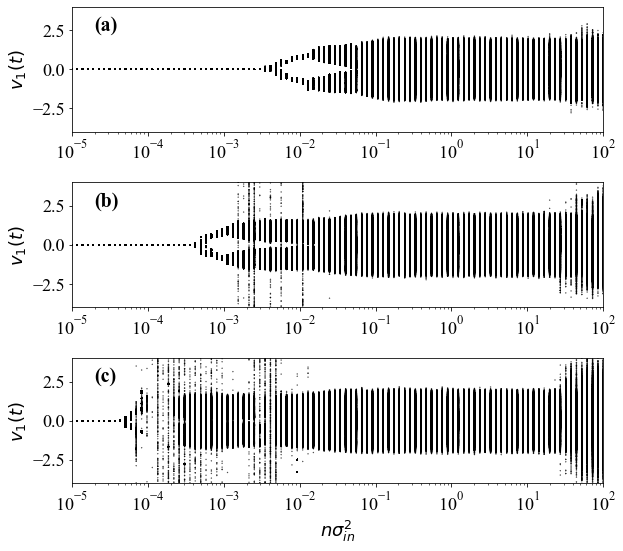}
    \caption{Bifurcation diagram of first component of the output $\mathbf{v}(t)$ of an ESN trained on the Lorenz63 time series with $sN\sigma_A^2=10^{-2}$. The ridge parameters are \textbf{(a)} $k=10^{0}$, \textbf{(b)} $k=10^{-2}$, and \textbf{(c)} $k=10^{-4}$.}
    \label{fig:bifurcation}
\end{figure}

When $n\sigma_{in}^2$ is small, prediction fails the full information case (see panels \textbf{(a)} and \textbf{(c)} in figure \ref{fig:heatmap}). To see what causes this, consider that in order for the ESN to predict a time series, it must be able to reproduce the Lyapunov spectrum of the input time series \cite{pathak2017using}. This means that the norm of the matrix $\mathbf{A}+\mathbf{W}^{(\rm in)}\mathbf{W}^{(\rm out)}$ relevant for the prediction dynamics \eref{eq:test_dyn}, must be sufficiently large. However, the ridge parameter $k$ sets a limit for how large the norm of $\mathbf{W}^{(\rm out)}$ can be. Consider, for example, a chaotic time series. To predict the chaotic time series, $n\sigma_{in}^2$ must exceed a threshold value so that the prediction dynamics can be chaotic. The same line of arguments hold for the case when partial information is provided (panels \textbf{(b)} and \textbf{(d)}). To observe the effect of changing the ridge parameter, we compute a bifurcation diagram of the reservoir neurons in an ESN trained on the Lorenz63 system. In figure \ref{fig:bifurcation}, we see how the ridge parameter changes at what value of $n\sigma_{in}^2$ the prediction dynamics bifurcates from having a stable fixed point at zero. Beyond this bifurcation, the prediction dynamics eventually becomes that of the Lorenz63 system. For smaller ridge parameters, the dynamics is more prone to become unstable. Indeed, the effect of the ridge parameter is to regularize $\mathbf{W}^{(\rm out)}$ such that its entries do not diverge to infinity due to $\mathbf{RR}^{\top}$ having an undefined inverse (see \eref{eq:ridge}). Thus, this instability is expected as $k$ decreases. The bifurcation is shown in figure \ref{fig:heatmap} as a blue line and corresponds to the second panel in figure \ref{fig:bifurcation}. In figure \ref{fig:heatmap}, the contour where the bifurcation occurs looks different for the full and partial information case because, for the case when only partial information is provided, the reservoir fails to embed the input dynamics and the prediction dynamics does not become chaotic.
\subsection{Independence on $\delta t$}\label{sec:sampling_rate}
\begin{figure}
    \centering
    \includegraphics[scale=0.6]{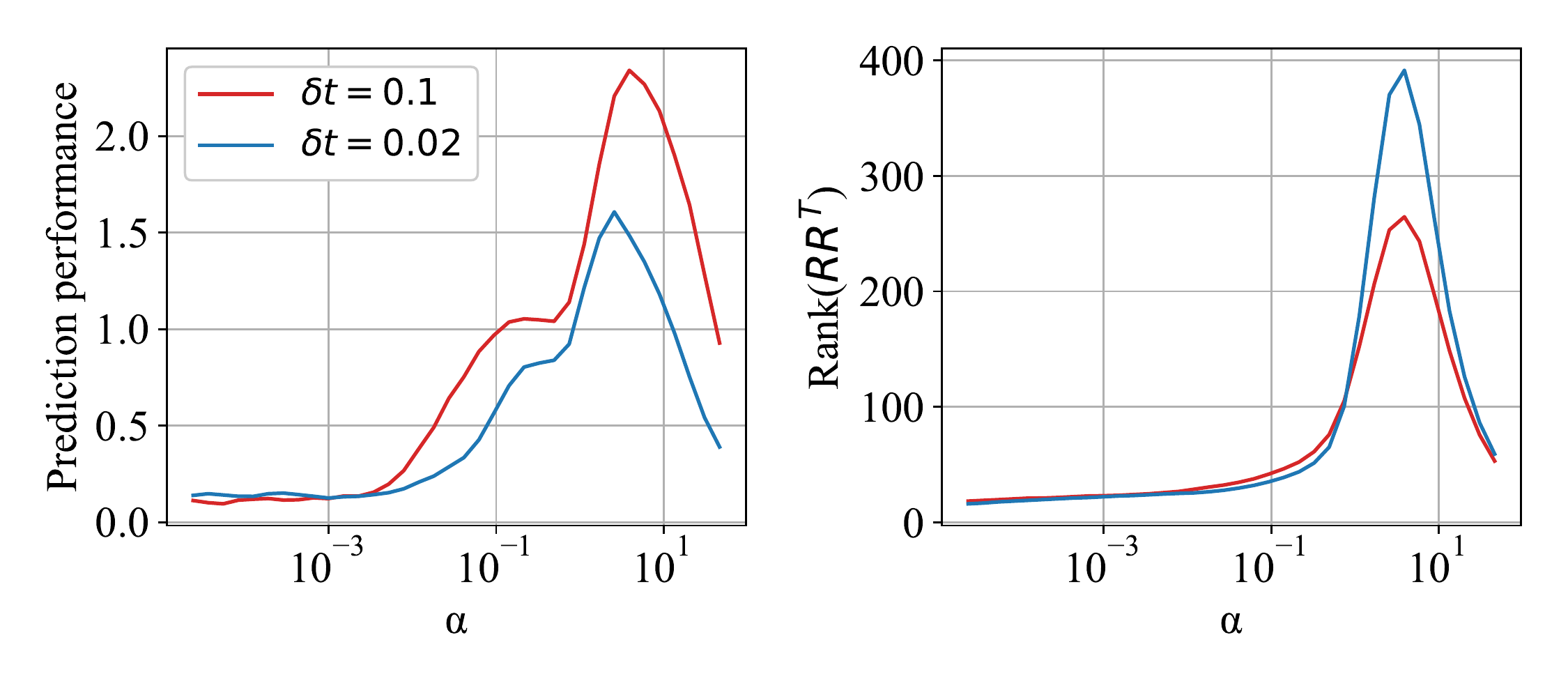}
    \caption{Prediction performance and $\mathbf{RR}^\top$ of two simple ESNs trained on the $y$-component of the Lorenz63 system, sampled at different rates $\delta t$. The reservoir dimension is $N=500$, and the result was averaged over 200 independent trials.}
    \label{fig:different_dt}
\end{figure}

To study the dependence on changing $\delta t$, we employ the ``simple ESN" architecture \cite{fette2005short}, where $\mathbf{A}$ is a diagonal matrix. This is done because it allows us to control the time scale of the reservoir neurons explicitly. In the result below, we deterministically set the diagonal elements of $\mathbf{A}$  to $A_{ii}=\alpha\frac{i}{N}$ for a positive parameter $\alpha$. The time scale of each neuron is simply determined by the magnitude of its corresponding weight in $\mathbf{A}$. If the ESN depends on $\delta t$, and by extension, the memory requirements of the time series to be predicted, the parameter region where prediction works should change when the sampling rate $\delta t$ is changed. 
As seen in figure \ref{fig:different_dt}, apart from decreasing the performance, decreasing $\delta t$ does not shift the parameter region where prediction works significantly, despite being altered by one order of magnitude. This is consistent with the previous observation, that the performance depends on the number of uncorrelated reservoir states, as measured by the rank of $\mathbf{RR}^\top$. What changes is instead the prediction performance. This is consistent with the result from \cite{kantz2004nonlinear}, where $\delta t=0.1$ is closer to the optimal sampling rate for time delay embedding of the Lorenz63 system. We note that the rank is larger when $\delta t$ is smaller.

\section{Conclusions}
Correctly selecting tuning parameters is crucial for successful reservoir computing. However, no clear understanding of how the parameters should be selected exists, and the choice largely comes down to heuristics. In this article, we explain how prediction performance depends on parameter selection when full phase space information or partial phase space information is provided to the network.

We find that there is a qualitative difference between the two cases. When partial phase space information is provided, the reservoir must construct a time-delay embedding of the input time series. To approximate the functional relationship between the embedding and the original space of the time series, the reservoir network uses a weighted sum of reservoir states; the more states, the more accurate the approximation. We show that the effective number of available reservoir states used for the approximation is equal to the number of independent states, quantifies by the number of non-zero singular values of the matrix $\mathbf{RR}^\top$. This imposes a condition on the relationship between the strength of the recurrent connections of the reservoir and the strength of the input signal. If the input signal dominates the dynamics, the reservoir states are strongly correlated, making the approximation of the functional relationship poor. On the other hand, no such condition is found when full phase space information is provided. This is because all the information required to predict the next time step is provided in the current time step. Hence, the reservoir network can simply perform function fitting to model the input time series.

That predictive performance improves when reservoir states become uncorrelated has a consequence for the role of memory capacity. As memory capacity increases when the correlation between the reservoir states at times $t$ and $t-\tau$ increases, maximizing memory capacity and predictive performance are mutually exclusive tasks. Memory capacity should therefore not be used as a metric associated with predictive performance.

Our results also show that tuning the time scale of the reservoir in accordance with the time scale of the input time series is unnecessary. In fact, the lower bound of the reservoir time scale for successful time-series prediction is independent on the sampling rate of the input time series. Instead, it depends on when the reservoir states start to become uncorrelated. However, we find that predictive performance can be improved by tuning the sampling rate in the same way it can be optimized in time-delay embedding literature.

Finally, we find that a lower limit for the strength of the input exists for both the full and partial information case due to that the ridge parameter limits the norm of the output connection strength. Limiting the norm constrains the maximum achievable maximal Lyapunov exponent of the reservoir dynamics during prediction. Hence, if this exponent is smaller than that of the input time series, prediction is impossible.

In conclusion, we have studied the parameter regions where reservoir computing is successful in the case of full and partial information, and found they differ qualitatively. The result is a step in the direction of clarifying how parameters should be selected in an informed way, instead of relying on heuristics. More research is needed to understand how the reservoir can be optimally designed to develop uncorrelated reservoir states to improve predictive performance. 
\mbox{}\\[3mm]
\noindent\textbf{Data availability statement}\\\noindent The data that supports the findings of this study are available upon reasonable request from the authors. 
\mbox{}\\[3mm]
\noindent\textbf{Ethical statement}\\\noindent This manuscript does not involve any human or animal participants.
\mbox{}\\[3mm]
\noindent\textbf{Conflict-of-interest statement}\\\noindent All authors declare that they have no conflicts of interest.
\vfill\eject

\appendix
\setcounter{section}{1}
\section*{Appendix}
The training Lyapunov exponent $\lambda_T$ is defined as
\begin{equation}
    \lambda_T=\lim_{t\rightarrow\infty}\frac{1}{t}\log\frac{|\mathbf{D}(t-1)\mathbf{A}\mathbf{D}(t-2)\mathbf{A}\dots\mathbf{D}(0)\mathbf{A}\delta \mathbf{r}(0)|}{|\delta\mathbf{r}(0)|},
    \label{eq:lya_appendix}
\end{equation}
where $\mathbf{D}(t)$ is a diagonal matrix with entries 
\begin{equation}
D_{ii}(t)=1-\tanh^2{\left(\sum_j^NA_{ij}r_j(t)+\sum_\alpha^n W_{i\alpha}^{(\rm in)}u_\alpha(t)\right)}\,,
\end{equation}
and $\delta r(t)$ is the separation between two initially infinitesimally nearby reservoir states. To derive \eref{eq:final_lya}, we start from \eref{eq:lya_appendix} by writing $\delta\mathbf{r}(0)=\delta r_0\mathbf{n}$, where $\mathbf{n}$ is the unit vector pointing in the direction of $\delta\mathbf{r}(0)$, and denote the matrix product as $\mathbf{J}_t=\mathbf{D}(t-1)\mathbf{A}\mathbf{D}(t-2)\mathbf{A}\dots\mathbf{D}(0)\mathbf{A}$. Using this, we write \eref{eq:lya} as
\begin{equation}
    \lambda_T=\lim_{t\rightarrow\infty}\frac{1}{2t}\ln{\left(\mathbf{n}^{\top}\mathbf{J}_t^{\top}\mathbf{J}_t\mathbf{n}\right)}.
\end{equation}
Assuming the decay of correlation between consecutive $\mathbf{D}(t)\mathbf{A}$ matrices is exponential, and that the distribution of the elements $D_{ii}(t)$ converge rapidly, we approximate the matrices $\mathbf{D}(t)\mathbf{A}$ as independent and identically distributed and use the Furstenberg theorem to obtain \cite{crisanti2012products}
\begin{equation}
    \lambda_T=\lim_{t\rightarrow\infty}\frac{1}{2t}\langle\ln{\left(\mathbf{n}^\top\mathbf{J}_t^\top\mathbf{J}_t\mathbf{n}\right)}\rangle,
\end{equation}
where the average is taken over samples of inputs and ensembles of $\mathbf{A}$ and $\mathbf{W}^{(\rm in)}$ matrices. We assume that the average over samples is equal to the time average of the input time series. The theorem states that in the limit of large $t$, the Lyapunov exponent is a non-random quantity. If the entries of $\mathbf{J}_t$ reach a stationary distribution, then the product $\mathbf{n}^\top\mathbf{J}_t^\top\mathbf{J}_t\mathbf{n}$ has a negligible variance in the limit of large $N$. In this limit, one obtains
\begin{equation}
    \lambda_T=\lim_{t\rightarrow\infty}\frac{1}{2t}\ln{\langle\mathbf{n}^\top\mathbf{J}_t^\top\mathbf{J}_t\mathbf{n}\rangle}.
\end{equation}
We use the result derived by Newman for products of i.i.d. random matrices \cite{newman1986distribution,crisanti2012products} to simplify the expression to
\begin{equation}
    \lambda_T=\frac{1}{2}\ln{\langle\mathbf{n}^\top(\mathbf{D}(t)\mathbf{A})^\top\mathbf{D}(t)\mathbf{A}\mathbf{n}\rangle}.
    \label{eq:mean_lambda}
\end{equation}
The proof of this equivalence requires the distribution of the random variable $\frac{|\mathbf{D}(t)\mathbf{A}\mathbf{z}(t)|}{|\mathbf{z}(t)|}$, where $\mathbf{z}(t)$ is a random $N$-dimensional vector, to be independent on $\mathbf{z}(t)$. Using the Euclidian norm, we have
\begin{equation}
    \frac{|\mathbf{D}(t)\mathbf{A}\mathbf{z}(t)|}{|\mathbf{z}(t)|}=\frac{\mathbf{z}^\top(t)\mathbf{A}^\top\mathbf{D}^2(t)\mathbf{A}\mathbf{z}(t)}{\mathbf{z}^\top(t)\mathbf{z}(t)}.
\end{equation}
The elements of the matrix $\mathbf{A}^\top\mathbf{D}^2(t)\mathbf{A}$ are sums of all the diagonal entries of $\mathbf{D}^2(t)$, each weighted by the product of two entries of $\mathbf{A}$. As the elements of $\mathbf{A}$ are i.i.d., when $N$ is large, this sum approaches a mean value that is independent of the direction of $\mathbf{z}(t)$. The proof then proceeds by stating that, if the random variable $\frac{|\mathbf{D}(t)\mathbf{A}\mathbf{z}(t)|}{|\mathbf{z}(t)|}$ is independent on $\mathbf{z}(t)$, then
\begin{equation}
    \ln|\mathbf{J}_t\mathbf{z}(0)|=\sum_{k=0}^{t-1}\ln\frac{|\mathbf{D}(k)\mathbf{A}\mathbf{z}(k)|}{|\mathbf{z}(k)|}
\end{equation}
is a sum of uncorrelated variables. The result in \eref{eq:mean_lambda} follows by employing the law of large numbers. Proceeding by using the assumption that the entries of $\mathbf{D}(t)\mathbf{A}$ are approximately i.i.d., \eref{eq:mean_lambda} can be evaluated to be
\begin{equation}
    \lambda_T=\frac{1}{2}\ln{N^{-1}\langle \textrm{tr}\left[(\mathbf{D}(t)\mathbf{A})^\top\mathbf{D}(t)\mathbf{A}\right]\rangle}.
\end{equation}
The argument of the logarithm can be rewritten as
\begin{eqnarray}
    N^{-1}\langle\textrm{tr}\left[\mathbf{A}^\top\mathbf{D}^2(t)\mathbf{A}\right]\rangle\nonumber
    &=N^{-1}\sum_i^N\left\langle D_{ii}^2(t)\left(\sum_j^NA_{ij}^2\right)\right\rangle\\
    &=N^{-1}\sum_i^N\left\langle D_{ii}^2(t)sN\sigma_A^2\right\rangle=s\sigma_A^2\sum_i^N\langle D_{ii}^2(t)\rangle.
\end{eqnarray}
Thus, we finally obtain
\begin{equation}
\label{eq:A10}
    \lambda_T=\frac{1}{2}\left[\ln{\left(sN\sigma_A^2\right)}+\ln{\left(N^{-1}\sum_i^N\langle D_{ii}^2(t)\rangle\right)}\right].
\end{equation}
This result is equivalent to (10) in  \cite{massar2013mean}, derived there for Gaussian white-noise inputs. Our derivation shows that \eref{eq:A10}  is valid for general, stationary time series with rapid decay of time correlations.
\mbox{}\\[3mm]

\providecommand{\newblock}{}

\end{document}